\documentclass[journal,onecolumn]{IEEEtran}
\usepackage[obeyFinal]{easy-todo}
\usepackage{soul}

\ifCLASSINFOpdf
\else
\fi
\usepackage{graphicx}
\usepackage{tikz}
\usetikzlibrary{circuits,circuits.logic}
\usetikzlibrary{circuits.logic.US}
\usepackage{tabularx}
\usepackage{float}

\begin{document}
% \title{Fully-parallel Hardware Implementation of Convolutional Neural Networks \\ for Edge Computing Applications}
%\title{Fully-parallel Convolutional Neural Network \\ Exploiting Stochastic Computing Correlation}
\title{Fully-parallel Convolutional \\ Neural Network Hardware}

\author{Christiam F. Frasser$^{\dagger}$, Pablo Linares-Serrano$^{\dagger\dagger}$, V. Canals$^{\dagger *}$, Miquel Roca$^{\dagger *}$, T. Serrano-Gotarredona$^{\dagger\dagger}$ \\and Josep L. Rossell\'{o}$^{\dagger *}$% <-this % stops a space
\thanks{$\dagger$ Electronics Engineering Group at Department of Physics, University of Balearic Islands, Ctra. Valldemossa Km 7.5, Palma de Mallorca 07122, Spain. 

$*$ Balearic Islands Health Research Institute, Palma de Mallorca, Spain.

$\dagger \dagger$ Instituto de Microelectr\'{o}nica de Sevilla (IMSE-CNM), CSIC, Seville, Spain

E-mail of corresponding author: \{j.rossello@uib.es\}}% <-this % stops a space
\thanks{This work has been partially supported by the Spanish Ministry of Science and innovation and the Regional European Development Funds (FEDER) under grant contracts TEC2017-84877-R and TEC2015-63884-C2-1-P. Pablo Linares-Serrano was supported by a JAE Intro ICUs 2019 scholarship at IMSE from the Spanish Research Council.}.% <-this % stops a space
\thanks{Manuscript received XXXXX; revised XXXX}}

\markboth{Journal of \LaTeX\ Class Files,~Vol.~XX, No.~XX, June~2020}%
{Shell \MakeLowercase{\textit{et al.}}: Bare Demo of IEEEtran.cls for IEEE Journals}

\maketitle
%-------------------------------------------------------------------
% 
%   ABSTRACT
%
%-------------------------------------------------------------------
\begin{abstract}
A new trans-disciplinary knowledge area, Edge Artificial Intelligence or Edge Intelligence, is beginning to receive a tremendous amount of interest from the machine learning community due to the ever increasing popularization of the Internet of Things (IoT). Unfortunately, the incorporation of AI characteristics to edge computing devices presents the drawbacks of being power and area hungry for typical machine learning techniques such as Convolutional Neural Networks (CNN). In this work, we propose a new power-and-area-efficient architecture for implementing Artiﬁcial Neural Networks (ANNs) in hardware, based on the exploitation of correlation phenomenon in Stochastic Computing (SC) systems. The architecture purposed can solve the difficult implementation challenges that SC presents for CNN applications, such as the high resources used in binary-to-stochastic conversion, the inaccuracy produced by undesired correlation between signals, and the stochastic maximum function implementation. Compared with traditional binary logic implementations, experimental results showed an improvement of 19.6x and 6.3x in terms of speed performance and energy efficiency, for the FPGA implementation. We have also realized a full VLSI implementation of the proposed SC-CNN architecture demonstrating that our optimization achieve a 18x area reduction over previous SC-DNN architecture VLSI implementation in a comparable technological node. For the first time, a fully-parallel CNN as LENET-5 is embedded and tested in a single FPGA, showing the benefits of using stochastic computing for embedded applications, in contrast to traditional binary logic implementations.
\end{abstract}

\IEEEpeerreviewmaketitle

%-------------------------------------------------------------------
% 
%   INTRODUCTION
%
%-------------------------------------------------------------------
\section{Introduction}

\IEEEPARstart{E}{dge} computing (EC) is characterized by the implementation of data processing at the edge of the network \cite{Shi2016637} instead of at the server level. This has produced great interest in the Microelectronic industry due to the proliferation of the Internet of Things (IoT). 
At the same time, incorporating Artificial Intelligence (AI) capacities in everyday devices has been in the spotlight in recent times, and it continues to be a hot topic, making the development of new techniques to extend AI to edge applications a must \cite{Zhou2019}. 
The idea behind these research efforts is to assist EC devices to further reduce their dependence on cloud processing by considerably reducing the energy associated with data transmission considering only the relevant information exchange with the cloud server.  
However, research on Edge Intelligence is still in its early days, since edge nodes normally present considerable limits in terms of area and power consumption, producing an intrinsic complexity for typical state-of-the-art deep learning implementations in embedded devices.
That is why new solutions for efficient hardware implementations for machine learning applications such as Convolutional Neural Networks (CNNs) have become a trending topic.

Stochastic Computing (SC), developed during the sixties \cite{gaines1969stochastic} as an alternative to traditional binary logic, is an approximate computing technique that has been arousing increasing interest over the last decade thanks to its capacity to compress complex functions within a low number of logic gates. 
Such characteristic has motivated the development of different proposals for the use of SC to implement ANNs in hardware
\cite{ren2017sc,8403283,7093194,Li2017NeuralNC,Rossello2010,Rossello2012}, and more specifically to implement CNN \cite{ren2017sc,8403283,sim2019cost,yu2017accurate},
facing the difficult SC implementation challenges such as: (a) the cost in terms of hardware resources required to implement different Random Number Generators (RNG), (b) the precision degradation between layers produced by the lack of full decorrelation between signals, and (c) the implementation of a stochastic maximum function circuitry.
Tackling these issues is not trivial, Lee et al. \cite{lee2017energy} has approached them by implementing only the first convolutional layer using stochastic computing. Sim et al. \cite{sim2017scalable} has created an hybrid stochastic-binary architecture, where only the multiplications are implemented in SC. Both approaches are not fully-stochastic, and therefore the benefits are limited.

In this work, we propose an efficient and compact hardware architecture to deal with these hurdles by exploiting  the correlation and decorrelation between SC signals in such a way as to implement the CNN basic building blocks. 
As a proof of concept, we implemented a fully-stochastic and parallel CNN on a single FPGA chip and compared its performance characteristics with different FPGA implementation works using traditional binary logic. This work is included in patent application \cite{josep2020}.

%-------------------------------------------------------------------
% 
%   2. STOCHASTIC COMPUTING
%
%-------------------------------------------------------------------
\section{Stochastic Computing}
%--------------------------------------
% 
%   2.1 UNIPOLAR AND BIPOLAR
%
%--------------------------------------
\subsection{Unipolar and bipolar codification}
Stochastic computing (SC) is an approximate computing methodology that represents signals using the switching frequency of time-dependent bit-streams. 
The SC signal is composed of pulses that represent the probability of finding a TRUE value (logic '1') at any arbitrary position throughout the sequence of bits. 
For instance, the number $0.75$ could be represented by a bit-stream in which the probability of finding a logic '1' along the bit-stream is $75\%$: (1,1,0,1) for a four bit-stream or (0,1,1,0,1,1,1,1) for an eight bit-stream. 
Representing only positive values (between 0 and 1) is known as \textit{unipolar} codification. 
To represent negative values, a different codification is required: the \textit{bipolar}  codification, where the number of zeros is subtracted from the number of ones, and finally divided by the total number of bits in the stream: $p^*=(N_1-N_0)/(N_0+N_1)$, where $N_0$ and $N_1$ are the number of zeros and ones respectively, and the $^*$ symbol denotes that the stochastic signal is represented in \textit{bipolar} codification. 
This expression is equivalent to implementing a change of variable: $p^*=2p-1$, where \textit{p} is the \textit{unipolar} representation of the number. As noted, the \textit{bipolar} codification provides a range of possible values in the interval $[-1,1]$. 

In the SC paradigm, each magnitude $X$ is converted to its time-dependent stochastic counterpart $x(t)$ by using a random number generator $R(t)$ and a comparator, so that the stochastic signal may be understood as a sequence of booleans $x(t)=\{X>R(t)\}$. 
If the number $X$ is greater than the random number $R(t)$, the output is set to '1' (assigned to the TRUE value of the comparison), otherwise it is set to '0' (FALSE value). 
If the random variable generated by $R(t)$ is uniform in the interval of all possible values of $X$, the mean switching probability $\bar{x}$ of the stochastic signal $x(t)$ is proportional to the converted magnitude $X$. 
In order to recover the $X$ value, a digital counter is incremented every high pulse of the bit-stream during a fixed period of time. 
The time length over which the sum is performed is related to the conversion error, so that the longer the time, the lower the error.

\begin{figure}[h]
	\centering
	\includegraphics[scale=1.0]{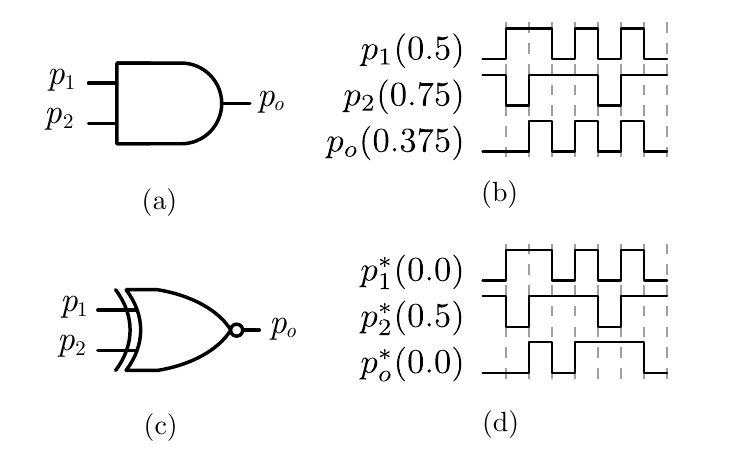}
	\caption{Difference in stochastic multiplication using different codification techniques with the same input bit-streams: (a) \textit{unipolar}  multiplication gate, (b) time diagram for \textit{unipolar} multiplication circuit, (c) \textit{bipolar} multiplication gate, (d) time diagram for \textit{bipolar} multiplication circuit.}
	\label{fig:stoch_mul}
\end{figure}

One of the main advantages of using SC is the low cost in hardware resources of implementing complex functions. 
Take for instance the multiplication operation, which is implemented in SC using just a single logic gate: an AND gate for \textit{unipolar} codification and an XNOR gate for \textit{bipolar} codification. Fig. \ref{fig:stoch_mul} shows how the same arithmetic operation could be achieved using different logic gates for different codification techniques in the presence of the same input waves.
%--------------------------------------------
% Grafica de stochastic multiplication
%--------------------------------------------

%--------------------------------------
% 
%   2.2 STOCHASTIC CORRELATION
%
%--------------------------------------
\subsection{Stochastic correlation}
\begin{figure}
\centering
\begin{tikzpicture}[circuit logic US,scale=0.9]
%
% Portes entrades
%
\node (Ai) at (0,0){$X$};
% BSC label
\node (labelbsc) at ($(Ai)+(-1,-2)$){$BSC$};
\draw [->] ($(labelbsc)+(0.5,0)$) -- ($(labelbsc)+(1.1,0.3)$);
\draw [->] ($(labelbsc)+(0.5,0)$) -- ($(labelbsc)+(1.1,-0.3)$);

% BSC Block X
\draw[fill=black!5,very thick,dashed] ($(Ai)+(0.2,0.6)$) -- ($(Ai)+(1.8,0.6)$)--($(Ai)+(1.8,-1.6)$) -- ($(Ai)+(0.2,-1.6)$)  --cycle;
%\draw[fill=black!5,very thick,dashed] ($(Ai)+(0.2,0.7)$) -- ($(Ai)+(1.8,0.7)$)--($(Ai)+(1.8,-1.5)$) -- ($(Ai)+(-0.6,-1.5)$)  -- ($(Ai)+(-0.6,-0.5)$) -- ($(Ai)+(0.2,-0.5)$) --cycle;
\node (Ri) at ($(Ai)+(-0.2,-1)$) {$R(t)$};
\draw[very thick] ($(Ai)+(0.2,0)$) -- ($(Ai)+(0.8,0)$);
\draw[very thick] ($(Ai)+(0.2,-1)$) -- ($(Ai)+(0.8,-1)$);
\draw[fill=white!20,very thick] ($(Ai)+(0.5,0.5)$) -- ($(Ai)+(1.5,0)$) -- ($(Ai)+(1.5,-1)$)  -- ($(Ai)+(0.5,-1.5)$) -- cycle;
\node (sml) at ($(Ai)+(1,-0.5)$) {$>$};
\node (asc) at ($(Ai)+(2.5,-0.5)$) {$x(t)$};

% BSC Block Y
\node (Ai2) at ($(Ai)+(0,-3)$){$Y$};
\draw[fill=black!5,very thick,dashed] ($(Ai2)+(0.2,0.6)$) -- ($(Ai2)+(1.8,0.6)$)--($(Ai2)+(1.8,-1.6)$) -- ($(Ai2)+(0.2,-1.6)$)  --cycle;
\draw[very thick] ($(Ai2)+(0.2,0)$) -- ($(Ai2)+(0.8,0)$);
\node (Ri2) at ($(Ai)+(-1,-4)$) {$R(t)\,or\, R'(t)$};
\draw[very thick] ($(Ai2)+(0.2,-1)$) -- ($(Ai2)+(0.8,-1)$);
\draw[fill=white!20,very thick] ($(Ai2)+(0.5,0.5)$) -- ($(Ai2)+(1.5,0)$) -- ($(Ai2)+(1.5,-1)$)  -- ($(Ai2)+(0.5,-1.5)$) -- cycle;
\node (sml2) at ($(Ai2)+(1,-0.5)$) {$>$};
\node (bsc) at ($(Ai2)+(2.5,-0.5)$) {$y(t)$};

\node [and gate,very thick,fill=white!40] (and) at ($(Ai)+(3,-2)$) {} ;

\draw[color=black] ($(Ai)+(1.5,-0.5)$) -- ++(right:6mm) |- (and.input 1);

\draw[color=black] ($(Ai2)+(1.5,-0.5)$) -- ++(right:6mm) |- (and.input 2);

\draw[color=black] (and.output) -- ++(right:1.5mm) ;
%\node (salida) at ($(and.output)+(1.6,0)$) {$\mathrm{min}(x(t),y(t))$ or $x(t)\cdot y(t)$};
\node (salida) at ($(and.output)+(1.6,0)$) {$\mathrm{min}(\bar{x},\bar{y})$};
\node (salida1) at ($(and.output)+(1.6,-0.4)$) {or};
\node (salida2) at ($(and.output)+(1.6,-0.8)$) {$\bar{x}\cdot \bar{y}$};
\end{tikzpicture}
\caption{Correlation impact over stochastic operations. Correlation between signals changes the operation computed by the logic gate. Stochastic signals $x(t)$ and $y(t)$ are said to be totally correlated when they share the same random number generator $R(t)$, producing the function $\mathrm{min}(\bar{x},\bar{y})$; otherwise, if $R'(t)$ is connected, they are said to be decorrelated and the output function is different: $\bar{x}\cdot \bar{y}$. 
\label{fig_and_gate_corr_uncorr}}
\end{figure}
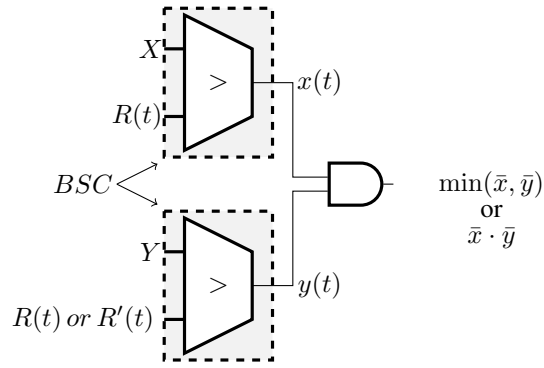

In order to generate stochastic bit-streams $x(t)$ from value $X$, a converter circuit must be implemented. The most commonly used circuit is based on a pseudo-random number generator, normally a Linear Feedback Shift Register (LFSR) and a comparator. 
The converter circuit is noted as BSC (Binary-to-Stochastic-Converter) in Fig. \ref{fig_and_gate_corr_uncorr}, where for the upper block the $X$ represents the signal to be converted, $R(t)$ the random value provided by the LFSR, and $x(t)$ the stochastic bit-stream generated. 

Two bit-streams are said to be correlated when both have some statistical similarities as discussed in \cite{6657023}. 
To produce the maximum correlation between two bit-streams, we can connect the same LFSR output $R(t)$ to the reference input of both comparators when performing the BSC conversion. 
A different pseudo-random number generator $R'(t)$ is used in case decorrelation is desired to operate the stochastic signals, producing a different outcome.

% Grafica de AND correlation 
To quantify the correlation, we can use the independence factor defined in \cite{7845709} or its dual, the stochastic computing correlation factor:
\begin{equation}
    C(x(t),y(t))=\frac{Cov(x(t),y(t))}{\mathrm{min}(\bar{x},\bar{y})-\bar{x}\bar{y}}
    \label{correlacion}
\end{equation}

where function $Cov$ is the covariance between the two time-dependent stochastic signals $x(t)$ and $y(t)$, while parameters $\bar{x}$, $\bar{y}$ are their mean values (their probability of being '1'). 
A correlation value of $+1$ implies maximum probabilistic similarity, obtained by sharing the same random number generator; whereas a $0$ value implies a complete decorrelation, produced by connecting two independent RNG as input reference comparison. 

The stochastic output of a two-inputs combinational gate can be expressed as a function of the correlation between its inputs. For the case of the AND and OR gates we have:
\begin{equation}
\begin{array}{rl}
    AND(x,y)= & \bar{x}\bar{y}+\bigl( \mathrm{min}(\bar{x},\bar{y})-\bar{x}\bar{y} \bigr) C(x,y)\\
    OR(x,y)= & \bar{x}+\bar{y}-\bar{x}\bar{y}+\bigl( \bar{x}\bar{y}-\mathrm{min}(\bar{x},\bar{y})\bigr) C(x,y)
\end{array}
\label{AND-OR}
\end{equation}

where as noted, the arithmetic operation is altered by the correlation level between the stochastic inputs.

Most of the errors produced by SC systems come from operating two stochastic signals with an undesired degree of correlation between them. 
Many works tries to operate with fully-uncorrelated stochastic signals, so they try to avoid the stochastic correlated imprecision by generating all the aleatory $R(t)$ signals with independent LFSRs, thus employing a high amount of hardware resources in the conversion circuits, and therefore limiting the contributions that SC offers for hardware implementations.
But despite the fact that decorrelation is necessary for many operations, there are some cases where correlated signals may be preferable. 
Consider the case of the AND gate of Fig. \ref{fig_and_gate_corr_uncorr}, in presence of two decorrelated signals (produced by using $R(t)$ and $R'(t)$ on each BSC), performs the multiplication operation: $\bar{x}\cdot \bar{y}$; while the \textit{minimum} operation $\mathrm{min}(\bar{x},\bar{y})$ is performed if the stochastic inputs are totally correlated (produced by sharing the same random generator $R(t)$ in the conversion circuit).
%, it performs the \textit{minimum} operation: $\mathrm{min}(x(t),y(t))$, if the stochastic inputs are totally correlated: $C(x(t),y(t))=1$ (produced by sharing the same random generator $R(t)$ in the conversion circuit).
This interesting feature can be exploited to produce high performance architectures that reduce area and power.

%--------------------------------------
% 
%   2.3 STOCHASTIC ADDITION
%
%--------------------------------------
\subsection{Stochastic addition}
\begin{figure}[h]
	\centering
	\includegraphics[scale=1.0]{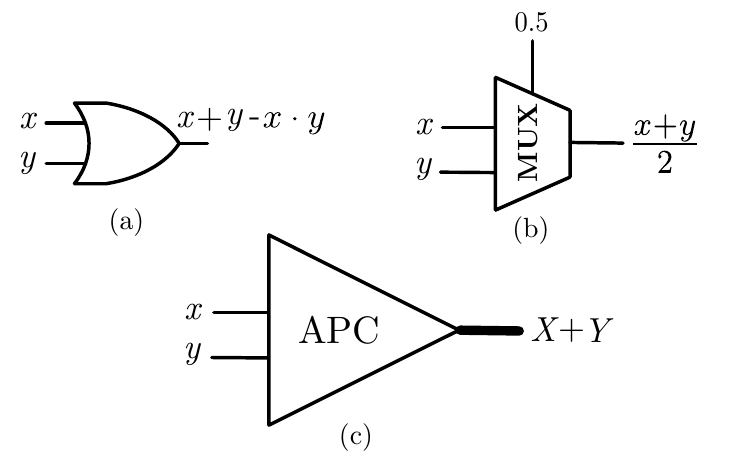}
	\caption{Stochastic addition circuits: (a) Stochastic addition using an OR gate, where $x\cdot y$ must be close to zero in order to compute the addition accurately. (b) Stochastic scaled addition using a multiplexer, where the accuracy outcome is dependant of the number of inputs. (c) Stochastic addition using an Accumulative Parallel Counter (APC), where the accuracy has no degradation and the output is represented in binary format.}
	\label{fig_add_sc}
\end{figure}

Due to boundaries of stochastic bit-stream representation, accurate implementations for the stochastic addition keeps being a challenge. Different circuits have been proposed to approximate the addition: a simple OR gate, a multiplexer, and an Accumulative Parallel Counter (APC). Fig. \ref{fig_add_sc} shows the different stochastic addition circuits, where for the sake of clarity, and from now on, the stochastic signals are denoted without the time-dependant reference $(t)$.

% Grafica de Addition SC

The OR gate is the smaller circuit in terms of hardware footprint, but it has the drawback of high inaccuracy outcomes when the input values are not close-to-zero enough. 
Not to mention its correlation dependant feature, discarding its use as an stochastic addition circuit in most of the applications.

The multiplexer is one of the most popular circuits to achieve the addition. 
The circuit is low-cost in terms of area and the precision is not affected by the correlation among the inputs.
The main disadvantage is the inaccuracy increment as the number of inputs grow, being not suitable for deep learning implementations, where the number of inputs for the addition operation are high demanding.

The last case is the APC, which counts the number of high pulses at the inputs and accumulates the counted value for a period of time, producing a complement-2 output. 
The APC solution is the most accurate from the circuits presented. Additionally, correlation among input signals does not disturb the result. 
The use of APC is the preferable approach for high precision implementations in spite of the higher resource utilization produced. 

%-------------------------------------------------------------------
% 
%   2.4 STOCHASTIC NEURON
%
%-------------------------------------------------------------------
\subsection{The Stochastic Neuron}

Convolutional Neural Networks (CNN) are constructed of several interconnected layers of neurons. 
The core neuron employed is composed of a scalar product block and a Rectified Linear Unit (ReLU) transfer function, which implements the operation: $\max(0,input)$.
The common base operations used in CNN implementations are: the multiplication, the addition and the maximum function, employed for the max-pooling operation and the ReLU transfer function. 
They can be easily implemented in stochastic computing systems if correlation is properly used. In the literature, different stochastic neuron designs have been proposed \cite{ren2017sc, 8403283, 7093194, Li2017NeuralNC}, although non of them have exploited the signal correlation properties, which can simplify the CNN hardware in a considerable way.

\begin{figure}
\centering
\begin{tikzpicture}[circuit logic US,scale=0.9]
%
% Portes entrades
%
\node [xnor gate,fill=white!40] (a1) at (0,0) {} ;
\node [xnor gate,fill=white!40] (a2) at (0,-1) {} ;
\draw[dotted] (0,-1.5) -- (0,-3.5) ;
\node [xnor gate,fill=white!40] (an) at (0,-4) {} ;
\draw[color=black] (a1.output) -- ++(right:3mm) |- (2,-2);
\draw[color=black] (a2.output) -- ++(right:3mm) |- (2,-2);
\draw[color=black] (an.output) -- ++(right:3mm) |- (2,-2);
\draw[color=black,ultra thick](0.9,-2) -- (2,-2);
\filldraw [black] (0.9,-2) circle [radius=2pt];
\node (bsz) at (1.6,-1.6){$n$}; 
\draw[color=black](1.5,-2.2) -- (1.7,-1.8);
%
%Labels Entrades neurona
%
\node (a) at (-1,0.11){};
\node (b) at (-1,-0.11){}; 
\node (as) at (-1.1,0.3){$x_{1}^*$};
\node (bs) at (-1.1,-0.3){$\omega_{i1}^*$}; 
\draw[color=black] (a) -- ++(right:3mm) -- (a1.input 1);
\draw[color=black] (b) -- ++(right:3mm) --(a1.input 2);
\node (ax) at (-1,-0.89){};
\node (bx) at (-1,-1.11){}; 
\node (asx) at (-1.1,-0.7){$x_{2}^*$};
\node (bsx) at (-1.1,-1.3){$\omega_{i2}^*$}; 
\draw[color=black] (ax) -- ++(right:3mm) -- (a2.input 1);
\draw[color=black] (bx) -- ++(right:3mm) --(a2.input 2);
\node (az) at (-1,-3.89){};
\node (bz) at (-1,-4.11){}; 
\node (asz) at (-1.1,-3.7){$x_{n}^*$};
\node (bsz) at (-1.1,-4.3){$\omega_{in}^*$}; 
\draw[color=black] (az) -- ++(right:3mm) -- (an.input 1);
\draw[color=black] (bz) -- ++(right:3mm) --(an.input 2);
%
% Blocs digitals
%
\draw[fill=white!20,very thick] (2,-3) -- (2,-1) -- (4,-2) -- cycle;
\node (apc) at (2.7,-2) {APC};
\draw[ultra thick,->](4,-2) -- (5,-2);

\draw[fill=white!20,very thick] (5,-2.4) -- (5,-1.6)--(6,-1.6) -- (6,-2.4)  -- cycle;
\node  at (5.5,-2) (A) {$BSC$};
\node at (5.5,-3.5) (Rt) {$R_x(t)$};
\draw [ultra thick,->](Rt) -- ($(Rt)+(0,1.1)$);
%\node (log2n) at (4.1,-1.1){$[1+log_2(N\cdot n)] $};
%\draw[->]($(log2n)+(0,-0.2)$) -- ($(log2n)+(0.4,-0.7)$);
%\draw(4.4,-2.2) -- (4.6,-1.8);
\draw[color=black](6,-2) -- (6.7,-2);
\node [or gate,fill=white!40] (a2) at (7,-2.2) {} ;
%\draw[color=black](6.45,-2.4) -- (6.7,-2.4);
%\node (rinp) at (6.2,-2.4){$0^*$};

\node (zero_sc) at (6.4,-3.3){$0^*$};
\node (rinp) at (6,-3){};
%\draw[color=black] (rinp) -- ++(right:3mm) |- (a2.input 2);
\draw[color=black] (zero_sc) |- (a2.input 2);

%\draw[fill=white!20,very thick] (7,-3) -- (7,-1.5) -- (7.7,-2)  -- (7.7,-2.5) -- cycle;
%\draw[ultra thick](6.5,-2.5) -- (7,-2.5);
%\node (lfsr) at (5.5,-2.8){$R(t)$}; 
%\node (lfsr) at (7.3,-2.25){$>$}; 
\draw[color=black](7.45,-2.2) -- (8,-2.2);
\node (sortida) at (8,-2.4){$y_{i}^*$}; 
\end{tikzpicture}
\caption{Stochastic neuron design exploiting correlation to reduce area cost. Stochastic signals from \textit{BSC} block and zero-\textit{bipolar} (\textit{$0^*$}) are generated with the same LFSR ($R_x(t)$) to produce total correlation between them, returning the maximum function on the output with a single OR gate.
\label{fig_neuron_sc}}
\end{figure}
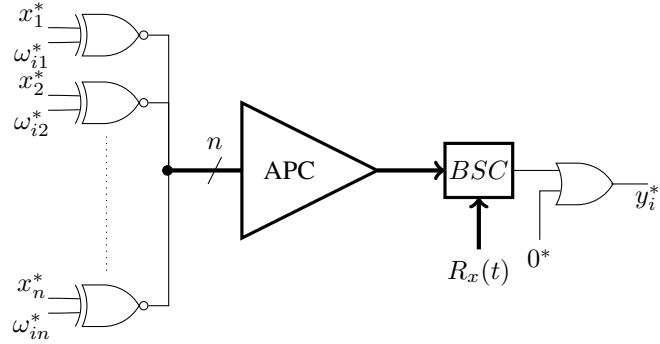

Fig. \ref{fig_neuron_sc} shows the proposed stochastic neuron design with the correlation exploiting architecture.
%The first block from the neuron is an XNOR gate array to compute the product between the inputs and weights.
%$x^{*}_{n}$ and weights $w_{in}$. 
%where i and n
%The addition is carried out by the APC block, which drives its binary output to the binary to stochastic converter (BSC), where once again, in stochastic domain, the OR gate operates as a ReLU transfer function.
The incoming stochastic vector $\mathbf{x^{*}}$ (composed of $n$ elements) is generated using the output  of one LFSR circuit $R_x(t)$, whereas the stochastic weight vector $\mathbf{w^*_{i}}$ (where $i$ represents the $i^{th}$ neuron of the current layer) is generated using the output  of a second LFSR circuit $R_w(t)$ (not shown in the diagram); therefore, producing decorrelation between them.
As a result, and considering a \textit{bipolar} codification, the $n$-XNOR-gate array calculates the stochastic product between neuron inputs and weights. 

Considering the APC is adding all the incoming signal products into one single complement-2 number, we can take advantage to connect the same pseudo-random number $R_x(t)$ used to generate 
%$\mathbf{x^{*}}$ and zero-\textit{bipolar} ($0^*$) signals, 
a zero-\textit{bipolar} ($0^*$) reference signal,
to the BSC block that re-converts the APC outcome to the stochastic domain, thus producing fully correlation between both stochastic signals.
Once in the stochastic domain, the ReLU activation function is easily implemented by using an OR gate, returning the operation: $y_i^*=\max(0^*,\sum_{j=1}^{n} x^*_j\cdot w^*_{ij})$.

Since the same procedure is followed in all of the neurons, two unique pseudo-random number generators ($R_x(t)$ and $R_w(t)$) are needed to accomplish the whole calculus, considerably saving area and power in the design.  

One of the benefits of the proposed stochastic-ReLU-function approach, is its normalized reproduction of the standard-ReLU-function used by the machine learning community. 
This means, that the weights obtained after the training process of the ANN, can be straight-forward adapted to the hardware, since  the expected activation function is not disturbing;
unlike other published studies, in which the function outcome is distorted, as is the case of references \cite{8403283,yu2017accurate}, where the stochastic implementation of the ReLU function, besides being large area-consuming, is clipped and not exact.
In the simple ReLU proposal, the OR gate implementation computes the maximum function without clipping or distorting the signal, and therefore, {\color{blue}the weights} of any standard training process considering ReLU-dependent neurons can be incorporated directly to the hardware after a simple process of normalization.

% Grafica de Neurona

%-------------------------------------------------------------------
% 
%   2.5 MAX POOLING
%
%-------------------------------------------------------------------
\subsection{Max-pooling layer}
In traditional Convolutional Neural Networks (CNN), after the convolutional layers extract the features from the input, a sub-sampling operation (pooling) is accomplished to reduce the spatial dimensions of the convoluted feature.
In the case of Max-Pooling (MP), the sub-sampling is performed selecting the maximum value from a spatial window of the convoluted feature: $\max_{j=1}^{k}(y^*_j)$, where $k$ is the size of the spatial window.

\begin{figure}[h]
	\centering
	\includegraphics[scale=1.0]{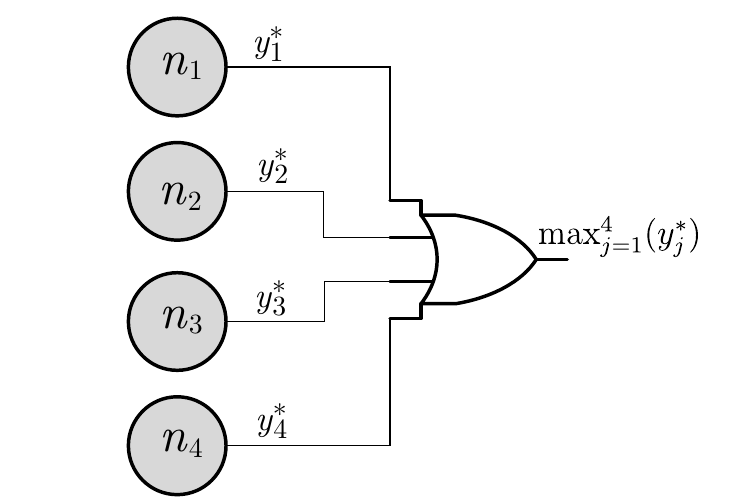}
	\caption{Stochastic max-pooling circuit	for a spatial window size of $k=2$x$2$. Stochastic neuron outputs $y^*_k$ are totally correlated, allowing the implementation of the maximum function with a single OR gate.}
	\label{fig_max_pool_sc}
\end{figure}

Ren et al.\cite{ren2017sc}, Z.Li et al.\cite{8403283}, and Yu et al.\cite{yu2017accurate} have proposed some stochastic maximum function designs using a set of counters, comparators and multiplexers.
%;employing a high utilization of hardware resources
The drawbacks of these architectures, are the employment of large area in hardware resources; and
moreover, the designs they proposed only find out the maximum value after counting the total number of high pulses in the bit-stream in a period of time, incurring in long latency and considerable energy consumption.
In contrast, our architecture takes advantage of the full correlation among the neuron outputs signals, and as in the ReLU transfer function case, it extracts the instantaneous maximum value with a unique OR gate (Fig. \ref{fig_max_pool_sc}), saving precious area resources, latency time and energy consumption. For the case of Min-pooling or average pooling, the OR gate must be changed by an AND gate or a multiplexer respectively.

% Grafica de Neurona

%=========================================================================
%   CNN Connection
%=========================================================================
\subsection{Full CNN Architecture}

\begin{figure*}[h]
	\centering
	\includegraphics[scale=0.45]{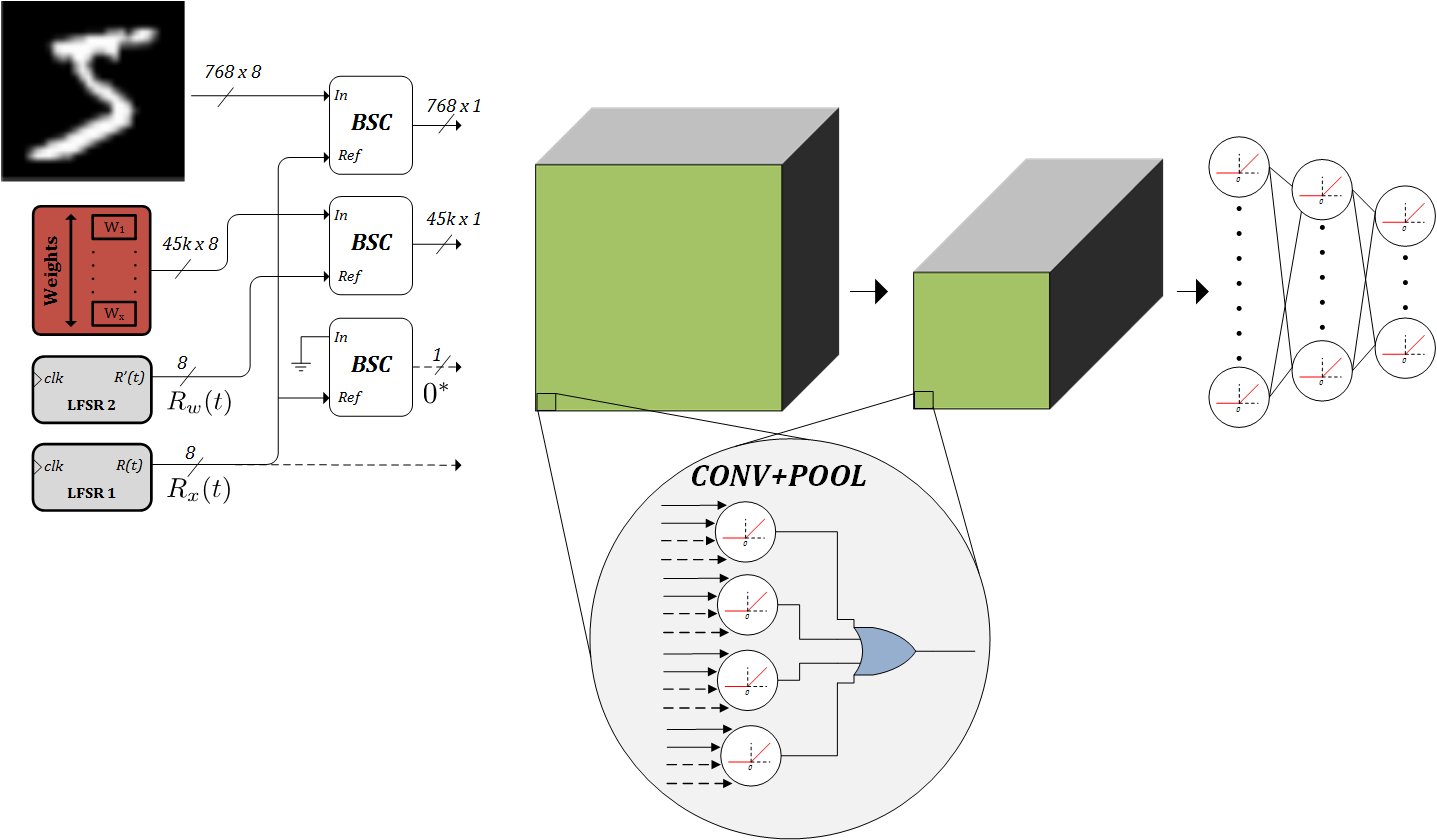}
	\caption{
	    {
	    Fully-parallel stochastic CNN architecture. Only two unique pseudo-random number generators are employed. All neurons are working simultaneously in parallel thanks to the correlation phenomenon exploitation. 
	    }
	}
	\label{fig_cnn_connection}
\end{figure*}

Figure \ref{fig_cnn_connection} shows how the whole system is connected in the CNN. 
As shown, only two unique pseudo-random number generators ($R_x(t)$ and $R_w(t)$) are needed to accomplish the whole calculus, considerably saving area and power in the design.  
This could be achieved thanks to the stochastic neuron design, which exploits correlation and de-correlation for computing.
LFSR1 is used for the $R_x(t)$ number generation, and is connected to the input image BSC conversion, the $0^*$ signal reference generator, and to every stochastic neuron in the whole design (for the APC stochastic generator, see Fig \ref{fig_neuron_sc}).
LFSR2 is used for the $R_w(t)$ number generation, which is only used to produce the stochastic weights. 
In this way, each stochastic signal generated by LFSR1 is totally uncorrelated with those generated by LFSR2, allowing neuron inputs to be multiplied by weights with the highest precision.
Moreover, the architecture proposed allows neuron outputs from layer $l_i$ to be connected to the neuron inputs of the next layer $l_{i+1}$ without any risk of signal degradation.
Since the $l_i$ neuron outputs are generated from a first LFSR block $R_x(t)$, and the $l_{i+1}$ weights are generated from a second LFSR $R_w(t)$, the error induced from layer to layer by the appearance of uncontrolled correlation between signals is totally avoided.

It is important to note that no prunning, weight sharing or clustering has been carried out in our architecture. The whole array of weights have been embedded in the design.

As noted by dashed lines, $R_x(t)$ and $0^*$ are shared through the whole network, saving plenty of resources and allowing all neurons work simultaneously in parallel. 
Power consumption is saved dramatically as no access to memory for reading or writing intermediate results is accomplished.

%-------------------------------------------------------------------
% 
%   EXPERIMENTS AND RESULTS
%
%-------------------------------------------------------------------
\section{Experimental Results}
In order to evaluate the proposed stochastic design, we have implemented the LeNet-5 Convolutional Neural Network (CNN). This CNN design is oriented to processing the MNIST handwriting data set composed of 60k training images and 10k testing images \cite{10027939599}. 
The CNN architecture consists of two convolutional layers and three fully connected layers, as the original paper from Lecun \cite{lecun1998gradient} describes. 
The baseline score of the trained model, using floating point, was $98.6\%$ (no special optimizations were introduced). 
On the other hand, the stochastic implementation score was $97.6\%$, only a $1\%$ accuracy degradation compared to the software version; a satisfactory result, considering no parameter fine tuning process was applied, just a simple weight normalization.
\subsection{FPGA Implementation}

We tested the full SC CNN implementation in a GIDEL PROC10A board, which has an Intel 10AX115H3F34I2SG FPGA running the 8-bit SC implementation at 150MHz. 
The communications were done through PCI express bus.

Table \ref{tab_comparison_lenet5} shows the comparison between the proposed implementation, with conventional FPGA-based CNN accelerators in terms of execution time latency, inference per second, and energy efficiency. 
%*****************************
% Tabla de comparaciones
%*****************************
\begin{table*}[ht]
    \centering
    \caption{
        Comparison with other FPGA Lenet-5 Implementations. 
    }
    \label{tab_comparison_lenet5}
    
    \begin{tabular}{lccccc}
    \hline
    
    \textbf{Model} & \textbf{FPGA16\cite{7544745}} & \textbf{FPGA17\cite{Liu2017ThroughputOptimizedFA}}  &
	\textbf{FPGA17\cite{8367260}}   &
	\textbf{FPGA18\cite{SoC2018Lenet5}}   &  
	\textbf{Proposed}\\ 
	\hline 
	
	Design method           &  Sequential (binary) 	& Sequential (binary) & Sequential (binary) & Sequential (binary) & Parallel (stochastic) \\	
	
	FPGA platform           &  Zynq XC7Z020 	& Virtex7 VX690T & Virtex7 485t & Zynq zc706  &  Arria10 GX1150 \\
	
	Frequency(MHz)  	    &  100	    & 100		& --        & 166       & 150   \\
	
	Latency(us) 	    	&  7916   	& 94.2  	& 960       & 1600      & 3.4   \\
	
	Kilo-Inferences per second (KIPS)&  0.13 & 10.6 & 1.04      & 0.625     & 294.1   \\
	
	Performance (KIPS/MHz) &  0.001     & 0.1       & --        & 0.003     & 1.96  \\
	
	Power(W)        	    &  --	    & 25.2		& 0.47		& 10.98     & 21 \\ 
	
	Energy efficiency(KI/J) &  --	    & 0.42		& 2.21 		& 0.056     & 14 \\ 
	
	Logic used (LUT/ALM)    & 9682      & 233K      & 7204      & 39837     & 343.4K \\
	
	Number of DSP used      & 2.4       & 2907      & 574       & 59        & 0 \\
	
	Memory Blocks used      & 6.16      & 477       & 343.3     & 97        & 0 \\
	
	Area efficiency (Inferences / MHz / ALM) & -- &--&--&--& 0.006\\
	\hline
    
    \end{tabular}
\end{table*}

As can be appreciated, the proposed method outperforms others architectures.
The results show that the proposed stochastic CNN implementation achieves 19.6x more speed performance (measured in inferences per second and per megahertz) compared to VX690T implementation \cite{Liu2017ThroughputOptimizedFA}, and a 6.3x more energy efficiency compared to  Virtex7-485t implementation \cite{8367260} (measured in inferences per Joule), making it promising for real embedded system applications. 

To the best of our knowledge, this is the first time an entire fully-parallel SC CNN is embedded into a single FPGA. This fact is in contrast to the studies presented in \cite{7544745,Liu2017ThroughputOptimizedFA,8367260, SoC2018Lenet5}, where the inference operations are accomplished using a loop-tiling technique (an optimization approach to use the same hardware resources recursively). 
%Thanks to the novel neuron design proposed, which reduces area consumption of the whole system, the proposed architecture fits an entire CNN into a single FPGA in a fully parallel implementation, unlike the studies in \cite{7544745,Liu2017ThroughputOptimizedFA,8367260, SoC2018Lenet5}, which iterates with the same logic resources recursively (loop-tiling technique). 
% The parallelization is the main reason for the lower latency of the proposed design, needing $2^N-1$ clock cycles (where $N$ is the bit resolution) to process the whole inference computation. 

%Talking about hardware resources needed per task, the proposed method stands out in relation to conventional FPGA implementations, where specific hardware blocks such as embedded memory and Digital Signal Processing blocks (DSPs) are used, 
In our design, DSPs blocks are avoided, since an unconventional computing technique (Stochastic Computing) is used instead of traditional binary logic. 
At the same time, memory blocks are not required, since the computation is not performed in a tile-loop manner, thus, reducing the main power consumption source, which comes from the access operations to the memory. 

In order to compare the area efficiency, the hardware area used by the FPGA specific blocks (DSP and memory) need to be known, but they are company-reserved; hence, we provide only the area efficiency value for our design (in terms of inferences per MHz and per logic unit ALM).
%Nonetheless, it is clear that the area occupied by the memory blocks plus DSP blocks seen in the reference studies, if implemented using only logic blocks, would exceed the logic resources occupied by our design.
%would be unsuitable even for the biggest FPGA in the market nowadays, reason enough to merit the proposed architecture, overall, if ASIC implementation is the target of the design. 

\subsection{VLSI Implementation}

The complete DNN stochastic architecture has also been synthesized in TSMC 40nm CMOS technology and in UMC 250 nm technology using the Cadence Genus Tool. The implemented design comprises a total number of 913,906 combinatorial elementary cells (NAND, NOR and inverter gates) and 104,317 sequential cells. 

The total area of the full design is 10.88 $mm^2$  in the UMC 250 nm technology node. This design obtains a 3x factor in area reduction with respect to a previously reported ASIC implementation in a 45nm technological node \cite{ren2017sc}. The design synthesized in the TSMC 40nm occupies a total area of 2.2 $mm^2$, which means that we get a 18x area reduction when using a comparable technological node. 
The main reason to achieve the area reduction is the compact implementation of the maximum function and max pooling operation by adequately exploiting the signal correlations. Furthermore, the use of correlated signals allows to implement the architecture using a very reduced number of pseudo-random number generators.

%-------------------------------------------------------------------
% 
%   CONCLUSIONS
%
%-------------------------------------------------------------------
\section{Conclusion}
Thanks to the advantages of the small area and low power consumption, stochastic computing is presented as a paradigm solution to implement machine learning algorithms in hardware for edge computing. However, many difficulties are still being faced in the quest to achieve good results. 
In this paper, we present an efficient reduced-area architecture to deal with the high area consumed by random number generators, the precision degradation produced by correlation between signals, and the stochastic maximum function implementation.
For the first time, a fully-parallel convolutional neural network is embedded in a single FPGA chip, obtaining better performance results compared to traditional binary logic implementations, showing the compression effectiveness of the architecture by exploiting the correlation features presented by stochastic signals.
The full parallel SC-CNN has been synthesized in a VLSI circuit demonstrating improved area efficiency over previous reported SC-DNN VLSI implementations.

\section{Author contributions}
C.F.F. conceived the experiments and performed the FPGA measurements, 
P.L.S. and T.S.G. implemented the VLSI part,
C.F.F. and J.L.R. conceived the design method.
Process supervised by V.C., M.R., T.S.G. and J.L.R.  
All authors
contributed to the discussion of the results and to the writing of the manuscript.
\bibliographystyle{IEEEtran}
\bibliography{main}

\end{document}